\pdfoutput=1

\documentclass[11pt]{article}

\usepackage[]{ACL2023}

\usepackage{times}
\usepackage{latexsym}

\usepackage[T1]{fontenc}

\usepackage[utf8]{inputenc}

\usepackage{microtype}

\usepackage{inconsolata}

\usepackage{amsmath}

\usepackage{multirow}
\usepackage{multicol}
\usepackage{makecell}
\usepackage{graphicx} 
\usepackage{amssymb}
\usepackage{algorithm}
\usepackage{algpseudocode}
\usepackage{ulem}
\usepackage{booktabs}
\usepackage{CJKutf8}
\usepackage{subfigure}

\usepackage{arydshln}
\usepackage{enumitem}

\usepackage{tikz}
\newcommand*{\circled}[1]{\lower.7ex\hbox{\tikz\draw (0pt, 0pt)%
    circle (.5em) node {\makebox[1em][c]{\small #1}};}}

\title{Improving Zero-shot Multilingual Neural Machine Translation by Leveraging Cross-lingual Consistency Regularization}

\author{Pengzhi Gao, Liwen Zhang, Zhongjun He, Hua Wu, and Haifeng Wang \\
Baidu Inc. No. 10, Shangdi 10th Street, Beijing, 100085, China \\
\texttt{\{gaopengzhi,zhangliwen04,hezhongjun,wu\_hua,wanghaifeng\}@baidu.com} 
}

\begin{document}

\maketitle

\begin{abstract}

The multilingual neural machine translation (NMT) model has a promising capability of zero-shot translation, where it could directly translate between language pairs unseen during training. For good transfer performance from supervised directions to zero-shot directions, the multilingual NMT model is expected to learn universal representations across different languages. This paper introduces a cross-lingual consistency regularization, CrossConST, to bridge the representation gap among different languages and boost zero-shot translation performance. The theoretical analysis shows that CrossConST implicitly maximizes the probability distribution for zero-shot translation, and the experimental results on both low-resource and high-resource benchmarks show that CrossConST consistently improves the translation performance\footnote{Source code: https://github.com/gpengzhi/CrossConST-MT}. The experimental analysis also proves that CrossConST could close the sentence representation gap and better align the representation space. Given the universality and simplicity of CrossConST, we believe it can serve as a strong baseline for future multilingual NMT research.

\end{abstract}

\section{Introduction}

\begin{figure}[h]
\centering
\includegraphics[scale=0.35]{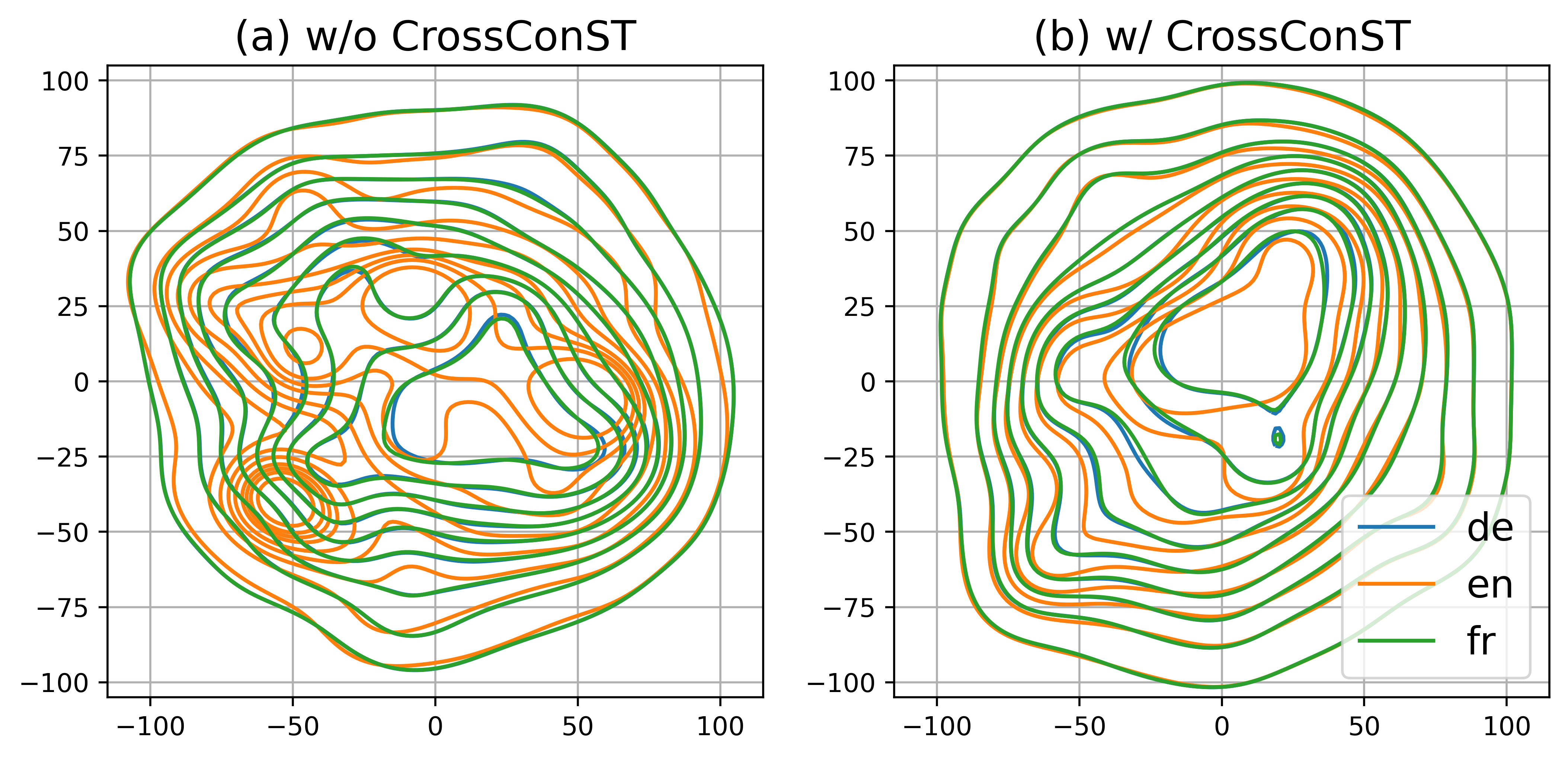}
\caption{Bivariate kernel density estimation plots of sentence representations after using T-SNE dimensionality reduction on the multi-way parallel testset \texttt{newstest2012}, where the max-pooled outputs of the multilingual NMT encoder are applied as the sentence representations. The blue line denotes Germany, the orange line denotes English, and the green line denotes French. This figure shows that the sentence representations are aligned better after utilizing CrossConST.}
\label{fig:kde}
\end{figure}

The objective of multilingual neural machine translation (NMT) is to construct a single, comprehensive model capable of translating between any pair of languages \cite{firat-etal-2016-multi,ha-etal-2016-toward,gu-etal-2018-universal,zhang-etal-2020-improving,JMLR:v22:20-1307}. 
This not only benefits low-resource translation \cite{aharoni-etal-2019-massively}, but also enables zero-shot translation \cite{gu-etal-2019-improved}. The success of zero-shot translation depends on the capability of the model to learn language-agnostic representations.
The conventional multilingual NMT model \cite{johnson-etal-2017-googles}, however, often struggles
with learning the universal representations among different languages (Figure \ref{fig:kde} (a)), which leads to poor zero-shot translation performance, particularly compared to the pivot-based methods \cite{ijcai2017p555}.


Several methods have been proposed to improve the zero-shot translation performance by learning language-agnostic representations and maximizing cross-lingual transfer. Some approaches modify the model architecture to achieve universal representations \cite{lu-etal-2018-neural,Ji_Zhang_Duan_Zhang_Chen_Luo_2020,liu-etal-2021-improving-zero,chen-etal-2021-zero}, 
while others utilize auxiliary training objectives to encourage similarity between the representations of different languages \cite{arivazhagan2019missing,al-shedivat-parikh-2019-consistency,pham-etal-2019-improving,pan-etal-2021-contrastive}. Specifically, \citet{gu-feng-2022-improving} introduce an agreement-based training approach to help the multilingual NMT model make consistent predictions based on the semantics-equivalent sentences. 
However, most existing methods are far from being widely used due to the degraded supervised translation performance, complicated algorithm implementation, and tedious hyperparameter search.

In this paper, our primary goal is to provide a simple, easy-to-reproduce, yet effective strategy for learning multilingual NMT. Inspired by \citet{gao-etal-2022-bi}, which boost the NMT performance by leveraging intra-lingual consistency regularization, we here propose a cross-lingual consistency regularization method, CrossConST, to learn the universal representations across different languages (Figure \ref{fig:kde} (b)) for boosting the zero-shot translation performance, where we introduce the explicit constraints to the semantic-equivalent sentence pairs by leveraging Kullback-Leibler (KL) regularization. The contributions of this paper can be summarized as follows:


\begin{itemize}

\item We propose CrossConST, a simple but effective method with only one hyperparameter for improving the generalization of the multilingual NMT model, and theoretically prove that it implicitly maximizes the probability distribution for zero-shot translation.

\item Our experimental results show that CrossConST achieves significant zero-shot translation improvements over the Transformer model on both low-resource and high-resource multilingual translation benchmarks and outperforms the state-of-the-art (SOTA) methods OT \& AT \cite{gu-feng-2022-improving} and mRASP2 \cite{pan-etal-2021-contrastive} on average.

\end{itemize}

\section{Cross-lingual Consistency for Multilingual NMT}

In this section, we formally propose CrossConST, a cross-lingual consistency regularization for learning multilingual NMT. 
We first review the multilingual neural machine translation (Section \ref{sec:mnmt}), then introduce our method in detail (Section \ref{sec:crossconst}). We theoretically analyze the regularization effect of CrossConST (Section \ref{sec:crossconst_analysis}) and propose a two-stage training strategy (Section \ref{sec:crossconst_training}).

\subsection{Multilingual Neural Machine Translation}\label{sec:mnmt}

Define $L = \{L_1, ..., L_M \}$, where $L$ is a collection of $M$ languages. The multilingual NMT model refers to a neural network with an encoder-decoder architecture, which receives a sentence in language $L_i$ as input and returns a corresponding translated sentence in language $L_j$ as output. Assume $\mathbf{x} = x_1, ..., x_I$ and $\mathbf{y} = y_1, ..., y_J$ that correspond to the source and target sentences with lengths $I$ and $J$, respectively. Note that $x_1$ denotes the language identification token to indicate the target language the multilingual NMT model should translate to, and $y_J$ denotes the special end-of-sentence symbol $\langle eos \rangle$. The encoder first maps a source sentence $\mathbf{x}$ into a sequence of word embeddings $e(\mathbf{x}) = e(x_1), ..., e(x_I)$, where $e(\mathbf{x}) \in \mathbb{R}^{d \times I}$, and $d$ is the embedding dimension. The word embeddings are then encoded to the corresponding hidden representations $\mathbf{h}$. Similarly, the decoder maps a shifted copy of the target sentence $\mathbf{y}$, i.e., $\langle bos \rangle, y_1, ..., y_{J-1}$, into a sequence of word embeddings $e(\mathbf{y}) = e(\langle bos \rangle), e(y_1), ..., e(y_{J-1})$, where $\langle bos \rangle$ denotes a special beginning-of-sentence symbol, and $e(\mathbf{y}) \in \mathbb{R}^{d \times J}$. The decoder then acts as a conditional language model that operates on the word embeddings $e(\mathbf{y})$ and the hidden representations $\mathbf{h}$ generated by the encoder.

Let $\mathcal{S}_{i,j}$ denote the parallel corpus of language pair $(L_i, L_j)$, and $\mathcal{S}$ denotes the entire training corpus. 
The standard training objective is to minimize the empirical risk:
\begin{equation}
\mathcal{L}_{ce}(\theta) =  \mathop{\mathbb{E}}\limits_{(\mathbf{x}, \mathbf{y}) \in \mathcal{S}} [\ell(f(\mathbf{x}, \mathbf{y}; \theta), \ddot{\mathbf{y}})],
\end{equation}
where $\ell$ denotes the cross-entropy loss, $\theta$ is a set of model parameters, $f(\mathbf{x}, \mathbf{y}; \theta)$ is a sequence of probability predictions, i.e., 
\begin{equation}
f_j(\mathbf{x}, \mathbf{y}; \theta) = P(y|\mathbf{x}, \mathbf{y}_{<j}; \theta),
\end{equation}
and $\ddot{\mathbf{y}}$ is a sequence of one-hot label vectors for $\mathbf{y}$.

\begin{figure*}[h]
\centering
\includegraphics[scale=0.5]{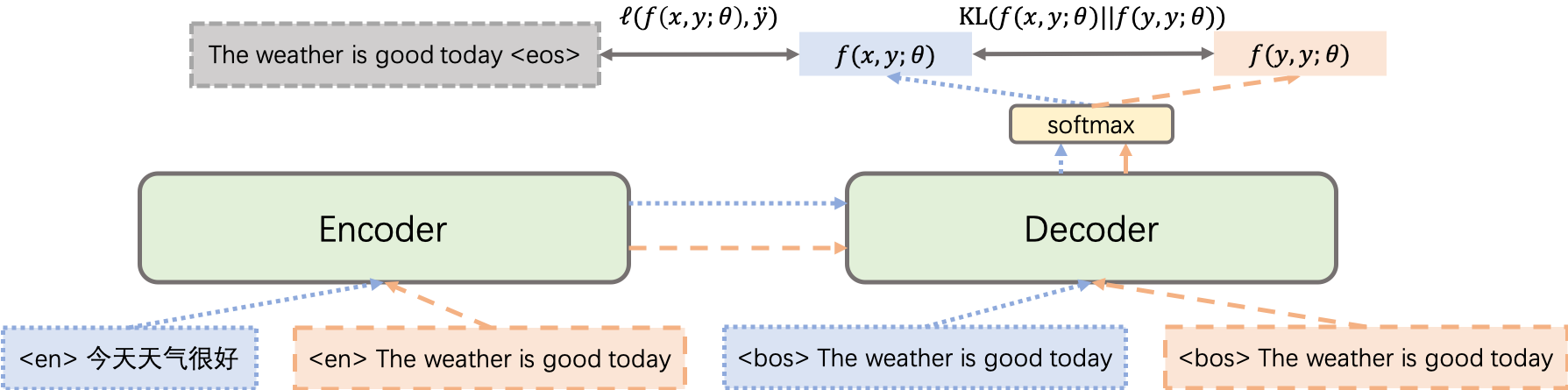}
\caption{Illustration of the CrossConST regularization, where the original Chinese-English sentence pair ("\begin{CJK}{UTF8}{gbsn}今天天气很好\end{CJK}", "The weather is good today") and the copied English-English sentence pair ("The weather is good today", "The weather is good today") are both go through the multilingual NMT model and obtain two output distributions $f(\mathbf{x}, \mathbf{y}; \theta)$ and $f(\mathbf{y}, \mathbf{y}; \theta)$. The same procedure is also applied to the English-Chinese sentence pair ("The weather is good today", "\begin{CJK}{UTF8}{gbsn}今天天气很好\end{CJK}") during the training of the multilingual NMT model.}
\label{fig:crossconst}
\end{figure*}

\subsection{CrossConST: A Cross-lingual Consistency Regularization for Multilingual NMT}\label{sec:crossconst}

Consider the multilingual NMT model as a function $f(\mathbf{x}, \mathbf{y}; \theta)$, which could be further decomposed as follows:
\begin{equation}
f(\mathbf{x}, \mathbf{y}; \theta) := f_{dec}(f_{enc}(\mathbf{x}; {\theta}_{enc}), \mathbf{y}; {\theta}_{dec}),
\end{equation}
where $f_{enc}(\cdot)$ and $f_{dec}(\cdot)$ denote the encoder and decoder, and ${\theta}_{enc}$ and ${\theta}_{dec}$ are the sets of parameters for the encoder and decoder respectively. An ideal multilingual NMT model should have the following properties:
\begin{itemize}
\item The encoder should output universal representations which are language agnostic. Semantic-equivalent sentences in different languages should share similar representations in the encoder output.

\item Given the target language to which the multilingual NMT model should translate to, the decoder should make consistent predictions based on the semantic-equivalent representations in the encoder output.
\end{itemize}

The main idea of our method is to close the representation gap among semantic-equivalent sentences in the encoder output and force the output distribution of the decoder to be consistent among different semantic-equivalent representations. During the training of multilingual NMT model, for each sentence pair $(\mathbf{x}, \mathbf{y})$, the training objective of CrossConST is defined as:
\begin{equation}\label{main_loss}
\mathcal{L}_{CrossConST}(\theta) = \mathcal{L}_{ce}(\theta) + \alpha \mathcal{L}_{kl}(\theta),
\end{equation}
where
\begin{equation}\label{kl_constraint}
\mathcal{L}_{kl}(\theta) = \text{KL}(f(\mathbf{x}, \mathbf{y}; \theta) \| f(\mathbf{y}, \mathbf{y}; \theta)),
\end{equation}
$\text{KL}(\cdot \| \cdot)$ denotes the Kullback-Leibler (KL) divergence of two distributions, and $\alpha$ is a scalar hyper-parameter that balances $\mathcal{L}_{ce}(\theta)$ and $\mathcal{L}_{kl}(\theta)$. Note that the gradient could be backpropagated through both sides of the KL regularization in CrossConST. Figure \ref{fig:crossconst} illustrates CrossConST regularization for learning multilingual NMT model.

Note that the constraint introduced by \eqref{kl_constraint} forces the equivalence between $f(\mathbf{x}, \mathbf{y}; {\theta})$ and $f(\mathbf{y}, \mathbf{y}; {\theta})$, which implicitly leads to
\begin{equation}
f_{enc}(\mathbf{x}; {\theta}_{enc}) = f_{enc}(\mathbf{y}; {\theta}_{enc}).
\end{equation}
Semantic-equivalent sentences $\mathbf{x}$ and $\mathbf{y}$ then share similar representations in the encoder output, and the decoder makes consistent predictions based on the semantic-equivalent representations $f_{enc}(\mathbf{x}; {\theta}_{enc})$ and $f_{enc}(\mathbf{y}; {\theta}_{enc})$. The properties of the ideal multilingual NMT model implicitly hold.

\subsection{Theoretical Analysis}\label{sec:crossconst_analysis}

Consider training a multilingual NMT model on the English-centric dataset, where $\mathbf{x}$ and $\mathbf{y}$ denote the sentences in two non-English languages, and $\mathbf{z}$ denotes the English sentence.
Let's consider the zero-shot translation direction $\mathbf{x}\rightarrow\mathbf{y}$. 
Inspired by \citet{ren-etal-2018-triangular} and \citet{wang-etal-2021-rethinking-zero}, we take a different approach to modeling the translation probability $P(\mathbf{y} | \mathbf{x}; \theta)$. We introduce language $\mathbf{z}$ as a bridge to connect $\mathbf{x}$ and $\mathbf{y}$. Following Jensen's Inequality, we could derive the lower bound of $P(\mathbf{y} | \mathbf{x}; \theta)$ over the parallel corpus $\mathcal{S}$ as follows:
\begin{align*}
\mathcal{L}(\theta) = & \sum_{(\mathbf{x}, \mathbf{y}) \in \mathcal{S}} \log P(\mathbf{y} | \mathbf{x}; \theta) \\
\ge & \sum_{(\mathbf{x}, \mathbf{y}) \in \mathcal{S}} \sum_{\mathbf{z}} Q(\mathbf{z}; \theta) \log \frac{P(\mathbf{y} | \mathbf{z}; \theta) P(\mathbf{z} | \mathbf{x}; \theta)}{Q(\mathbf{z}; \theta)} \\
:= & \bar{\mathcal{L}}(\theta),
\end{align*}
and the gap between $\mathcal{L}(\theta)$ and $\bar{\mathcal{L}}(\theta)$ could be calculated as follows:
\begin{align*}
\mathcal{L}(\theta) - \bar{\mathcal{L}}(\theta) = & \sum_{(\mathbf{x}, \mathbf{y}) \in \mathcal{S}} \sum_{\mathbf{z}} Q(\mathbf{z}; \theta) \log \frac{Q(\mathbf{z}; \theta)}{P(\mathbf{z} | \mathbf{y}; \theta)} \\
= & \sum_{(\mathbf{x}, \mathbf{y}) \in \mathcal{S}} \text{KL}(Q(\mathbf{z}; \theta) \| P(\mathbf{z} | \mathbf{y}; \theta)),
\end{align*}
where $Q(\mathbf{z}; \theta)$ is an arbitrary posterior distribution of $\mathbf{z}$. Note that we utilize the approximation that $P(\mathbf{y} | \mathbf{x}, \mathbf{z}; \theta) \approx P(\mathbf{y} | \mathbf{z}; \theta)$ and $P(\mathbf{z} | \mathbf{x}, \mathbf{y}; \theta) \approx P(\mathbf{z} | \mathbf{y}; \theta)$ due to the semantic equivalence of parallel sentences $\mathbf{x}$ and $\mathbf{y}$.

We then introduce the autoencoding task of $\mathbf{z}$ by replacing $Q(\mathbf{z}; \theta)$ with $P(\mathbf{z} | \mathbf{z}; \theta)$ such that
\begin{align}\label{eq:lower}
\bar{\mathcal{L}}(\theta) = & \sum_{(\mathbf{x}, \mathbf{y}) \in \mathcal{S}} \mathop{\mathbb{E}}_{\mathbf{z} \sim P(\mathbf{z} | \mathbf{z}; \theta)} \log P(\mathbf{y} | \mathbf{z}; \theta) \nonumber \\
& - \text{KL}(P(\mathbf{z} | \mathbf{z}; \theta)\|P(\mathbf{z} | \mathbf{x}; \theta))
\end{align}
and
\begin{align}\label{eq:gap}
\mathcal{L}(\theta) - \bar{\mathcal{L}}(\theta) = \sum_{(\mathbf{x}, \mathbf{y}) \in \mathcal{S}} \text{KL}(P(\mathbf{z} | \mathbf{z}; \theta) \| P(\mathbf{z} | \mathbf{y}; \theta)).
\end{align}
To maximize $\mathcal{L}(\theta)$, we should maximize the lower bound $\bar{\mathcal{L}}(\theta)$ and minimize the the gap between $\mathcal{L}(\theta)$ and $\bar{\mathcal{L}}(\theta)$. By utilizing the cross-lingual consistency regularization, CrossConST helps minimize the KL terms in \eqref{eq:lower} and \eqref{eq:gap} and implicitly maximizes the probability distributions for zero-shot translation, which results in better translation performance in $\mathbf{x}\rightarrow\mathbf{y}$ direction. The detailed proof can be found in Appendix \ref{sec:crossconst_theory}.

\subsection{Training Strategy: Multilingual NMT Pretraining and CrossConST Finetuning}\label{sec:crossconst_training}


Inspired by \citet{johnson-etal-2017-googles} and \citet{wu-etal-2021-language}, we only use one language tag to indicate the target language the multilingual NMT model should translate to. For instance, the following English to German sentence pair ``How are you? $\rightarrow$ Wie geht es dir?'' is transformed to ``<\texttt{de}> How are you? $\rightarrow$ Wie geht es dir?''. And \citet{wu-etal-2021-language} demonstrate that such language tag strategy could enhance the consistency of semantic representations and alleviate the off-target issue in zero-shot translation directions.

To stabilize the multilingual NMT training procedure and accelerate the convergence of the multilingual NMT model, we adopt a two-stage training strategy. We first train a conventional multilingual NMT model as the pretrained model and then finetune the model with CrossConST objective function \eqref{main_loss}. It is worth mentioning that \citet{pham-etal-2019-improving} derive a similar problem formulation and training strategy. However, they do not demonstrate the effectiveness of their proposed method (KL Softmax) in \citet{pham-etal-2019-improving}. To the best of our knowledge, we for the first time show the effectiveness of the simple cross-lingual consistency regularization for improving the translation performance of the multilingual NMT model. Note that while \citet{pham-etal-2019-improving} decouple the gradient path in the decoder from the KL divergence term, our design allows for backpropagation through both sides of the KL regularization in CrossConST. We do not decouple any gradient path in our model.

\section{Low Resource Scenario}\label{sec:low-resource}

We here investigate the performance of CrossConST on the low-resource multilingual machine translation benchmark. For fair comparisons, we keep our experimental settings consistent with the previous work \cite{gu-feng-2022-improving}.

\subsection{Dataset Description}

We conduct our experiments on the IWSLT17 benchmark \cite{cettolo-etal-2017-overview}, which releases a multilingual corpus in five languages: English (\texttt{en}), German (\texttt{de}), Dutch (\texttt{nl}), Romanian (\texttt{ro}), and Italian (\texttt{it}). We consider the English-centric scenario, where we collect the parallel sentences from/to English. The detailed information of the training dataset is summarized in Table \ref{iwslt17_statics} in Appendix \ref{sec:appendix_dataset}. There are eight supervised translation directions and twelve zero-shot translation directions, and we use the official validation and test sets in our experiments. Following the common practice, we tokenize each language by applying the Moses toolkit \cite{koehn-etal-2007-moses} and build a shared dictionary with 32K byte-pair-encoding (BPE) \cite{sennrich-etal-2016-neural} types.

\subsection{Model Configuration}

We implement our approach on top of the Transformer \cite{vaswani2017attention}. We apply a standard base Transformer with 6 encoder and decoder layers, 8 attention heads, embedding size 512, and FFN layer dimension 2048. We apply cross-entropy loss with label smoothing rate $0.1$ and set max tokens per batch to be $4096$. We use  the Adam optimizer with Beta $(0.9, 0.98)$, $4000$ warmup updates, and inverse square root learning rate scheduler with initial learning rates $7e^{-4}$. We use dropout rate $0.3$ and beam search decoding with beam size $5$ and length penalty $0.6$. We apply the same training configurations in both pretraining and finetuning stages. We fix $\alpha$ to be $0.25$ in \eqref{main_loss} for CrossConST. We use case-sensitive sacreBLEU \cite{post-2018-call} to evaluate the translation quality. We train all models until convergence on eight NVIDIA Tesla V100 GPUs. All reported BLEU scores are from a single model. 
For all the experiments below, we select the saved model state with the best validation performance.

\subsection{Main Results}\label{main_results}

\begin{table*}[h]\small
\centering
\begin{tabular}{l | c c c c c c c | c} 
\multicolumn{1}{c|}{Method} & \multicolumn{1}{c}{\texttt{de} $\leftrightarrow$ \texttt{it}} & \multicolumn{1}{c}{\texttt{de} $\leftrightarrow$ \texttt{nl}} & \multicolumn{1}{c}{\texttt{de} $\leftrightarrow$ \texttt{ro}} & \multicolumn{1}{c}{\texttt{it} $\leftrightarrow$ \texttt{ro}} & \multicolumn{1}{c}{\texttt{it} $\leftrightarrow$ \texttt{nl}} & \multicolumn{1}{c}{\texttt{nl} $\leftrightarrow$ \texttt{ro}} & Zero-shot & Supervised \\
& & & & & & & Average & Average \\
\hline
\hline
Pivot$^{\dagger}$ & 18.10 & 19.66 & 16.49 & 21.37 & 21.44 & 18.70 & 19.29 & - \\
\hdashline
m-Transformer$^{\dagger}$ & 15.46 & 18.30 & 14.70 & 19.03 & 18.48 & 16.11 & 17.01 & 30.62 \\
SR Alignment$^{\dagger}$ & 16.45 & 18.80 & 15.45 & 20.02 & 19.20 & 17.25 & 17.85 & 30.41 \\
KL-Softmax$^{\dagger}$ & 16.06 & 18.27 & 15.00 & 20.09 & 18.89 & 16.52 & 17.46 & 30.50 \\
mRASP2 w/o AA$^{\dagger}$ & 16.98 & 19.60 & 15.88 & 20.75 & 19.40 & 17.59 & 18.36 & 30.39 \\
DisPos$^{\dagger}$ & 16.13 & 19.21 & 15.52 & 20.12 & 19.58 & 17.32 & 17.97 & 30.49 \\
DAE Training$^{\dagger}$ & 16.32 & 18.69 & 15.72 & 20.42 & 19.11 & 17.22 & 17.91 & 30.51 \\
TGP$^{\dagger}$ & 17.64 & 15.85 & 16.86 & 19.34 & 19.53 & 20.05 & 18.21 & 30.66 \\
LM Pretraining$^{\dagger}$ & 17.66 & 15.86 & 16.16 & 19.05 & 19.02 & \bf 20.07 & 17.96 & 30.52 \\
OT \& AT$^{\dagger}$ & 17.28 & 19.81 & 16.09 & 20.83 & 20.14 & 17.85 & 18.66 & 30.52 \\
\hline
Pivot & 18.87 & 20.09 & 17.20 & 21.56 & 22.22 & 19.35 & 19.88 & - \\
\hdashline
OT \& AT & 18.18 & 20.22 & 16.82 & 21.96 & 21.15 & 18.66 & 19.50 & 31.14 \\ 
m-Transformer & 17.2 & 19.61 & 15.88 & 20.81 & 20.21 & 17.89 & 18.60 & 31.34 \\
\ \ + CrossConST & \bf 18.70 & \bf 20.32 & \bf 16.98 & \bf 22.17 & \bf 21.83 & 19.30 & \bf 19.88 &  \bf 31.37
\end{tabular}
\caption{Performance on the IWSLT17 multilingual translation benchmark. Each entry in the first six columns denotes the averaged BLEU scores of both directions. $\dagger$ denotes the numbers are reported from \citet{gu-feng-2022-improving}, others are based on our runs. The highest scores are marked in bold for all models except for the pivot translation in each column. The detailed evaluation results are summarized in Table \ref{iwslt17-eval} in Appendix \ref{sec:iwslt17_eval}.}
\label{iwslt17_results}
\end{table*}

We compare our approach with the following methods on the IWSLT17 benchmark:

\begin{itemize}[leftmargin=*]

\item {\bf m-Transformer} \cite{johnson-etal-2017-googles}: A multilingual NMT model that directly learns the many-to-many translation on the English-centric dataset.

\item {\bf Pivot Translation} \cite{ijcai2017p555}: m-Transformer first translates the source language into English before generating the target language.

\item {\bf Sentence Representation Alignment (SR Alignment)} \cite{arivazhagan2019missing}: An additional regularization loss is utilized to minimize the discrepancy of the source and target sentence representations.

\item {\bf Softmax Forcing (KL-Softmax)} \cite{pham-etal-2019-improving}: This method forces the decoder to generate the target sentence from itself by introducing a KL divergence loss.

\item {\bf Contrastive Learning (mRASP2 w/o AA)} \cite{pan-etal-2021-contrastive}: This method introduces a contrastive loss to minimize the representation gap between the similar sentences and maximize that between the irrelevant sentences. Note that the aligned augmentation (AA) method is not utilized.

\item {\bf Disentangling Positional Information (DisPos)} \cite{liu-etal-2021-improving-zero}: This method drops the residual connections in a middle layer of the encoder to achieve the language-agnostic representations.

\item {\bf Denosing Training (DAE Training)} \cite{wang-etal-2021-rethinking-zero}: This approach introduces a denoising autoencoding task during the multilingual NMT model training.

\item {\bf Target Gradient Projection (TGP)} \cite{yang-etal-2021-improving-multilingual}: This method guides the training with constructed oracle data, where the gradient is projected not to conflict with the oracle gradient.

\item {\bf Language Model Pretraining (LM Pretraining)} \cite{gu-etal-2019-improved}: This approach strengthens the decoder language model (LM) prior to NMT model training.

\item {\bf Optimal Transport \& Agreement-based Training (OT \& AT)} \cite{gu-feng-2022-improving}: This method proposes an optimal transport loss to bridge the gap between the semantic-equivalent representations and an agreement-based loss to force the decoder to make consistent predictions based on semantic-equivalent sentences. We set $\gamma_1$ and $\gamma_2$ in OT \& AT to be $0.4$ and $0.001$ respectively in the experiments.

\end{itemize}

We report test BLEU scores of all comparison methods and our approach on the IWSLT17 dataset in Table \ref{iwslt17_results}. We can see that our multilingual NMT model achieves strong or SOTA BLEU scores in both supervised and zero-shot translation directions. Note that our approach outperforms OT \& AT even though its implementation is much more complicated than ours. It is worth mentioning that CrossConST is the only method that can achieve a similar zero-shot translation performance compared with the pivot translation. Note that the BLEU scores of our m-Transformer, especially in the zero-shot translation directions, are higher than that reported in \citet{gu-feng-2022-improving}. Such gap might be due to the different language tag strategies used in \citet{gu-feng-2022-improving} and our experiments, which is in line with \citet{wu-etal-2021-language}.

\subsection{Does CrossConST Still Work Beyond English-centric Scenario?}

We here extend our experiments on the IWSLT17 benchmark beyond the English-centric scenario. Specifically, we gather the English-centric dataset used in Section \ref{main_results} and supplement it with an additional 20K \texttt{de} $\leftrightarrow$ \texttt{it} sentence pairs, which are subsampled from the IWSLT17 dataset. This experimental setup is highly practical because the size of the non-English datasets is usually an order less than that of the English-centric dataset.

\begin{table}[h]\small
\centering
\begin{tabular}{l | c | c | c} 
\multicolumn{1}{c|}{Method} & Training & Zero-shot & Supervised \\
& Dataset & Average & Average \\
\hline
\hline
m-Transformer & \circled{1} & 18.60 & 31.34 \\
\ \ + CrossConST & \circled{1} & 19.88 & 31.37 \\
\hline
m-Transformer & \circled{2} & 19.76 & 31.59 \\
\ \ + CrossConST & \circled{2} & \bf 20.35 & \bf 31.67 \\
\end{tabular}
\caption{Performance on the IWSLT17 multilingual translation benchmark. \circled{1} denotes the English-centric dataset. \circled{2} denotes the English-centric dataset + extra \texttt{de} $\leftrightarrow$ \texttt{it} dataset. The detailed evaluation results are summarized in Table \ref{iwslt17-extra-eval} in Appendix \ref{sec:iwslt17_eval}.}
\label{iwslt17_extra_results}
\end{table}

We report test BLEU scores of the baseline and our approach on the IWSLT17 dataset in Table \ref{iwslt17_extra_results}. By checking model performance under different combinations of dataset and training strategy, we have the following observations: 1) Adding beyond the English-centric dataset (\texttt{de} $\leftrightarrow$ \texttt{it}) could greatly improve the overall zero-shot translation performance. 2) The CrossConST is complementary to the data-based method and could further improve the performance of the zero-shot translation.

\section{High Resource Scenario}\label{sec:high-resource}

We here investigate the performance of the CrossConST on the high-resource multilingual machine translation benchmark. For fair comparisons, we keep our experimental settings consistent with the previous works \cite{lin-etal-2020-pre,pan-etal-2021-contrastive}.

\begin{table*}[h]\small
\centering
\begin{tabular}{l | c | c c c c c c c c | c} 
\multicolumn{1}{c|}{Method} & Training & \multicolumn{2}{c}{\texttt{en} - \texttt{fr}} & \multicolumn{2}{c}{\texttt{en} - \texttt{tr}} & \multicolumn{2}{c}{\texttt{en} - \texttt{es}} & \multicolumn{2}{c|}{\texttt{en} - \texttt{fi}} & Average \\
& Dataset & \multicolumn{2}{c}{WMT14} & \multicolumn{2}{c}{WMT17} & \multicolumn{2}{c}{WMT13} & \multicolumn{2}{c|}{WMT17} \\
& & $\rightarrow$ & $\leftarrow$ & $\rightarrow$ & $\leftarrow$ & $\rightarrow$ & $\leftarrow$ & $\rightarrow$ & $\leftarrow$ \\
\hline
\hline
m-Transformer$^{\dagger}$ & \circled{1} & 42.0 & 38.1 & 18.8 & 23.1 & 32.8 & 33.7 & 20.0 & 28.2 & 29.66 \\
mRASP2 w/o AA$^{\dagger}$ & \circled{1} & 42.1 & 38.7 & 18.2 & 24.8 & 33.1 & 33.2 & 20.0 & 27.8 & 29.74 \\
mRASP$^{\dagger}$ & \circled{2} & 43.1 & 39.2 & 20.0 & 25.2 & 34.0 & 34.3 & 22.0 & 29.2 & 30.88 \\
mRASP2 w/o MC24$^{\dagger}$ & \circled{2} & 43.3 & 39.3 & 20.4 & \bf 25.7 & 34.1 & \bf 34.3 & 22.0 & 29.4 & 31.06 \\
\hline
mRASP2$^{\dagger}$ & \circled{3} & 43.5 & 39.3 & 21.4 & 25.8 & 34.5 & 35.0 & 23.4 & 30.1 & 31.63 \\
\hline
m-Transformer & \circled{1} & 43.5 & 40.3 & 20.8 & 23.8 & 33.4 & 32.7 & 22.0 & 28.8 & 30.66 \\
\ \ + CrossConST & \circled{1} & 44.1 & \bf 40.7 & 21.2 & 24.5 & 33.8 & 33.0 & 22.2 & 29.5 & 31.13 \\
mRASP & \circled{2} & 44.5 & 39.7 & 22.1 & 23.6 & 33.9 & 33.1 & 23.3 & 29.0 & 31.15 \\
\ \ + CrossConST & \circled{2} & \bf 44.6 & \bf 40.7 & \bf 22.4 & 24.4 & \bf 34.3 & 33.7 & \bf 23.5 & \bf 29.7 & \bf 31.66 \\ 
\end{tabular}
\caption{Performance (tokenized BLEU) on WMT supervised translation directions. $\dagger$ denotes the numbers are reported from \citet{pan-etal-2021-contrastive}, others are based on our runs. The highest scores are marked in bold for all models except for mRASP2 in each column. \circled{1} denotes PC32. \circled{2} denotes PC32 + RAS. \circled{3} denotes PC32 + RAS + MC24.}
\label{main_supervised}
\end{table*}

\subsection{Dataset Description}


We conduct our experiments on PC32, a multilingual parallel corpus of $32$ English-centric language pairs. 
We collect the pre-processed PC32 dataset from \citet{lin-etal-2020-pre}'s release\footnote{https://github.com/linzehui/mRASP}. 
We also collect the pre-processed PC32 dataset after applying random aligned substitution (RAS) technique from \citet{lin-etal-2020-pre}'s release. 
The detailed statistics of all training datasets are summarized in Tables \ref{pc32_statics} and \ref{pc32_ras_statics} in Appendix \ref{sec:appendix_dataset}.


For supervised directions, we collect testsets from WMT benchmarks, where four languages, Spanish (\texttt{es}), Finnish (\texttt{fi}), French (\texttt{fr}), and Turkish (\texttt{tr}), are selected, resulting in 8 translation directions. We use $\texttt{multi-bleu.pl}$\footnote{https://github.com/moses-smt/mosesdecoder/blob/ master/scripts/generic/multi-bleu.perl} for tokenized BLEU \cite{papineni-etal-2002-bleu} evaluation, where both reference and hypothesis are tokenized by Sacremoses\footnote{https://github.com/alvations/sacremoses}. For zero-shot directions, we collect OPUS-100 zero-shot testsets from \citet{zhang-etal-2020-improving}'s release\footnote{https://opus.nlpl.eu/opus-100.php}, where six languages, Arabic (\texttt{ar}), German (\texttt{de}), French (\texttt{fr}), Dutch (\texttt{nl}), Russian (\texttt{ru}), and Chinese (\texttt{zh}), are selected, resulting in 25 translation directions. Note that Dutch is not covered in our training dataset such that we only evaluate the zero-shot directions when Dutch is at the source side. We evaluate the multilingual NMT models by case-sensitive sacreBLEU \cite{post-2018-call}.

\subsection{Model Configuration}

We apply a Transformer with $12$ encoder and decoder layers, 16 attention heads, embedding size 1024, and FFN layer dimension 4096. We use dropout rate $0.1$, learning rate $3e^{-4}$ with polynomial decay scheduling and 10000 warmup updates. We use Adam optimizer with Beta $(0.9, 0.98)$ and $\epsilon=1e^{-6}$. We set the threshold of gradient norm to be $5.0$. We apply cross-entropy loss with label smoothing rate $0.1$ and set max tokens per batch to be $1536$ with update frequency $50$. We use beam search decoding with beam size $5$ and length penalty $1.0$. We apply the same training configurations in both pretraining and finetuning stages. We fix $\alpha$ to be $0.1$ in \eqref{main_loss} for CrossConST. We train all models until convergence on $8 \times 4$ NVIDIA Tesla V100 GPUs. All reported BLEU scores are from a single model. 
We select the saved model state with the best validation performance for all the experiments below.

\begin{table*}[h]\small
\centering
\begin{tabular}{l | c c c c c c c c c c c | c} 
\multicolumn{1}{c|}{Method} & \multicolumn{2}{c}{\texttt{x} - \texttt{ar}} & \multicolumn{2}{c}{\texttt{x} - \texttt{zh}} & \multicolumn{1}{c}{\texttt{x} - \texttt{nl}$^*$} & \multicolumn{2}{c}{\texttt{x} - \texttt{fr}} & \multicolumn{2}{c}{\texttt{x} - \texttt{de}} & \multicolumn{2}{c|}{\texttt{x} - \texttt{ru}} & Avg. \\
& $\rightarrow$ & $\leftarrow$ & $\rightarrow$ & $\leftarrow$ & $\leftarrow$ & $\rightarrow$ & $\leftarrow$ & $\rightarrow$ & $\leftarrow$ & $\rightarrow$ & $\leftarrow$ & \\
\hline
\hline
Pivot$^{\dagger}$ & 5.5 & 21.1 & 28.5 & 20.3 & 6.0 & 26.1 & 23.9 & 14.4 & 16.6 & 16.6 & 24.6 & 18.22 \\
\hdashline
m-Transformer$^{\dagger}$ & 3.7 & 6.7 & 6.7 & 5.0 & 6.3 & 7.7 & 5.0 & 4.2 & 4.9 & 5.7 & 5.6 & 5.60 \\
mRASP2 w/o AA$^{\dagger}$ & 4.8 & 17.1 & 26.1 & 15.8 & 6.4 & 22.9 & 21.2 & 11.8 & 15.3 & 13.3 & 21.4 & 15.79 \\
mRASP$^{\dagger}$ & 4.1 & 4.4 & 8.2 & 4.0 & 5.1 & 2.4 & 7.6 & 6.2 & 4.1 & 4.1 & 4.6 & 4.97 \\
mRASP2 w/o MC24$^{\dagger}$ & \bf 5.9 & \bf 18.3 & \bf 27.5 & 16.5 & \bf 9.6 & \bf 25.2 & 21.6 & 11.2 & \bf 16.7 & 15.6 & \bf 21.7 & 17.07 \\
\hline
mRASP2$^{\dagger}$ & 5.3 & 20.8 & 29.0 & 17.7 & 6.1 & 23.6 & 23.0 & 12.3 & 16.4 & 16.4 & 22.8 & 17.32 \\
\hline
Pivot (m-Transformer) & 6.6 & 22.2 & 29.5 & 21.4 & 8.7 & 27.5 & 24.7 & 15.7 & 17.1 & 18.0 & 25.3 & 19.46 \\
Pivot (mRASP) & 6.9 & 21.9 & 29.4 & 21.8 & 8.1 & 27.2 & 25.3 & 15.5 & 17.2 & 18.3 & 25.6 & 19.49 \\
\hdashline
m-Transformer & 5.3 & 11.2 & 17.4 & 16.5 & 7.5 & 16.8 & 21.3 & 9.8 & 13.1 & 14.5 & 8.2 & 12.75 \\
\ \ + CrossConST & 5.4 & 17.7 & 27.2 & 18.4 & 9.3 & 24.0 & 23.9 & 14.0 & 16.0 & 15.9 & 20.5 & 17.30 \\
mRASP & 5.6 & 13.7 & 24.1 & 18.3 & 7.2 & 17.7 & 23.0 & 11.1 & 13.1 & 15.5 & 15.5 & 14.80 \\
\ \ + CrossConST & \bf 5.9 & 16.7 & 27.2 & \bf 19.6 & 9.2 & 23.5 & \bf 24.6 & \bf 14.3 & 16.0 & \bf 16.4 & 20.9 & \bf 17.48 \\
\end{tabular}
\caption{Performance (de-tokenized BLEU using SacreBLEU) on OPUS-100 zero-shot translation directions. $\dagger$ denotes the numbers are reported from \citet{pan-etal-2021-contrastive}, others are based on our runs. $*$ indicates that Dutch (\texttt{nl}) is not included in PC32. The highest scores are marked in bold for all models except for the pivot translation and mRASP2 in each column. }
\label{main_zeroshot}
\end{table*}

\subsection{Main Results}

We compare our approach with the following methods on the PC32 benchmark:

\begin{itemize}[leftmargin=*]


\item {\bf mRASP} \cite{lin-etal-2020-pre}: This method proposes a random aligned substitution (RAS) technique that builds code-switched sentence pairs for multilingual pretraining. Note that the results of mRASP reported in this paper are obtained without finetuning.

\item {\bf mRASP2} \cite{pan-etal-2021-contrastive}: This method utilizes the RAS technique on both the bilingual dataset (PC32) and an additional monolingual dataset (MC24). It introduces a contrastive loss to minimize the representation gap between the similar sentences and maximize that between the irrelevant sentences. mRASP2 w/o AA only adopts the contrastive loss based on m-Transformer, and mRASP2 w/o MC24 excludes MC24 from mRASP2.

\end{itemize}



We report test BLEU scores of all comparison methods and our approach on WMT supervised translation directions in Table \ref{main_supervised}. With CrossConST regularization, our multilingual NMT model achieves strong or SOTA BLEU scores on the supervised translation directions. 
Note that all comparison methods and our approach share the same model architecture, and the only differences are the training dataset and the objective loss function.
We report test BLEU scores of all comparison methods and our approach on OPUS-100 zero-shot translation directions in Table \ref{main_zeroshot}, which includes six languages and 25 translation directions in total. The detailed evaluation results are summarized in Table \ref{main_zeroshot_details} in Appendix \ref{sec:zeroshot}. We also report the evaluation results of the pivot translation based on m-Transformer and mRASP. We can see that CrossConST greatly boosts the performance in the zero-shot translation directions and substantially narrows the performance gap with the pivot translation. It is worth mentioning that our approach could improve zero-shot translation by a large margin and also benefit the supervised translation.




By checking model performance under different scenarios, we have the following observations: 1) Our language tag strategy works better than that in \citet{pan-etal-2021-contrastive} for learning the multilingual NMT model on the English-centric dataset, especially for the zero-shot translation, which is in line with \citet{wu-etal-2021-language}. 2) CrossConST is crucial for the performance improvement in the zero-shot translation directions and performs slightly better when combined with the code-switched training dataset. 3) Our approach could outperform mRASP2 on average in the absence of the MC24 dataset, which implies the effectiveness of CrossConST compared with the contrastive loss utilized in mRASP2.

\subsection{Does CrossConST Really Learn A Better Latent Representation?}\label{sec:analysis}


We conduct the experiments on the multi-way parallel testset \texttt{newstest2012}\footnote{https://www.statmt.org/wmt13/dev.tgz} from the WMT13 \cite{bojar-etal-2013-findings} translation task, where 3003 sentences have translations in six languages: Czech (\texttt{cs}), Germany (\texttt{de}), English (\texttt{en}), Spanish (\texttt{es}), French (\texttt{fr}), and Russian (\texttt{ru}). We calculate the sentence representations by max-pooling the multilingual NMT encoder outputs.
\paragraph{Sentence Representation Visualization} To verify whether CrossConst can better align different languages’ semantic space, we visualize the sentence representations of Germany (\texttt{de}), English (\texttt{en}), and French (\texttt{fr}). We apply dimension reduction on the 1024-dimensional sentence representations with T-SNE \cite{NIPS2002_6150ccc6} and then depict the bivariate kernel density estimation based on the 2-dimensional representations in Figure \ref{fig:kde}. Figure~\ref{fig:kde} shows that m-Transformer cannot align these three languages well in the representation space, while CrossConST draws the sentence representations across different languages much closer. Please check Figures \ref{fig:kde_baseline} and \ref{fig:kde_crossconst} in Appendix \ref{sec:visualization} for the visualization of the sentence representations in other languages.

\begin{figure}[h]
\centering
\includegraphics[scale=0.55]{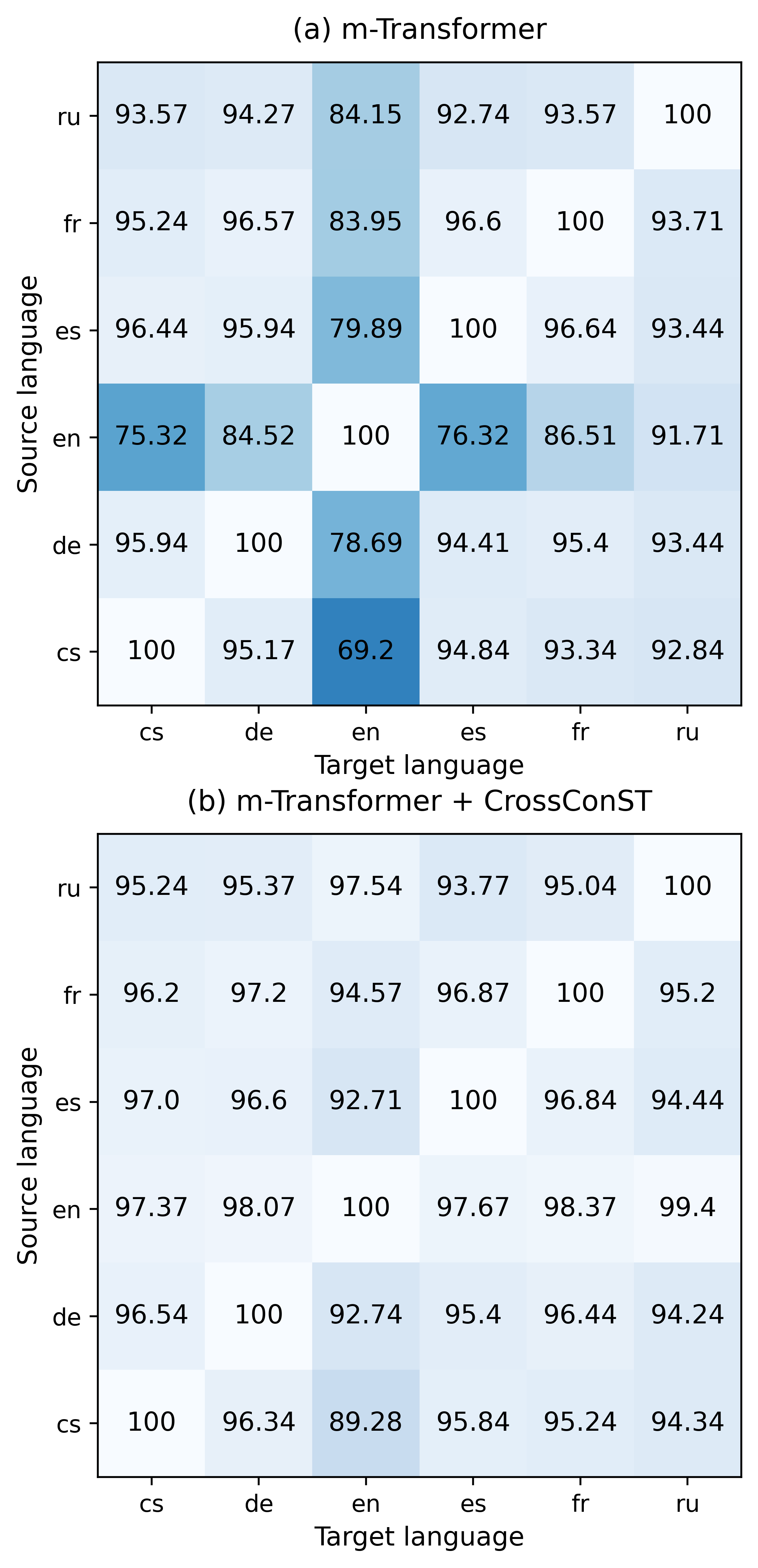}
\caption{Similarity search accuracy of m-Transformer with/without CrossConST for different language pairs.}
\label{fig: similarity_search}
\end{figure}

\paragraph{Multilingual Similarity Search}
We conduct the multilingual similarity search experiment to verify that CrossConST indeed closes the latent representation gap among different languages. For each sentence in the source language, we find the closest sentence in the target language according to the cosine similarity of the corresponding sentence representations. 
The evaluation results are reported in Figure \ref{fig: similarity_search}. By checking model performance on different language pairs, we have the following observations: 1) m-Transformer could achieve decent performance (94.71\% on average) among non-English directions. However, the similarity search accuracy degrades dramatically (81.03\% on average) in the English-centric directions, which implies that English does not align well with non-English languages in m-Transformer. We think such bad representation alignment between English and non-English languages is one of the critical reasons that m-Transformer underperforms in the zero-shot translation directions compared with the pivot-based method. 2) CrossConST significantly improves the similarity search performance in the English-centric direction (14.74\% improvement on average) and further boosts the performance among non-English directions (1\% improvement on average). We believe the improvement of similarity search accuracy could be regarded as an indicator of better cross-lingual representation alignment and confirm that CrossConST can learn effective universal representation across different languages.

\section{Related Work}

Early works on multilingual NMT demonstrate its zero-shot translation capability \cite{ha-etal-2016-toward,johnson-etal-2017-googles}. To further improve the zero-shot translation performance, one direction is to force the multilingual NMT encoder output to be language-agnostic via additional regularization constraints or training tasks \cite{pham-etal-2019-improving,arivazhagan2019missing,wei2020learning,liu-etal-2021-improving-zero,wang-etal-2021-rethinking-zero,yang-etal-2021-improving-multilingual,gu-feng-2022-improving}. For example, \citet{gu-feng-2022-improving} introduce an agreement-based training approach to help the multilingual NMT model make consistent predictions based on the semantics-equivalent sentences. Our method follows this line but outperforms these methods by introducing a simple yet effective cross-lingual regularization constraint, which effectively reduces discrepancies in representations across languages.

Another direction is to utilize extra data such as generated pseudo sentence pairs, monolingual datasets, and pretrained models \cite{gu-etal-2019-improved,al-shedivat-parikh-2019-consistency,zhang-etal-2020-improving,chen-etal-2021-zero,yang-etal-2021-multilingual}. For example, \citet{al-shedivat-parikh-2019-consistency} encourages the multilingual NMT model to produce equivalent translations of parallel training sentence pairs into an auxiliary language. \citet{zhang-etal-2020-improving} proposes random online back-translation to enforce the translation of unseen training language pairs. Unlike these methods, CrossConST does not require additional data and is orthogonal to these methods. We could further boost the zero-shot translation performance by combining our method with these data-driven approaches.




\section{Conclusion}\label{sec:conclusion}

In this paper, we propose CrossConST: a simple but effective cross-lingual consistency regularization method for learning multilingual NMT models. We theoretically analyze the regularization effect of CrossConST and verify its effectiveness for zero-shot translation. For the stable training of multilingual NMT, we propose a two-state training strategy that consists of multilingual NMT pretraining and CrossConST finetuning. Experiments on low and high resource multilingual translation benchmarks demonstrate CrossConST's capabilities to improve translation performance in both supervised and zero-shot directions. Further experimental analysis confirms that our method indeed leads to better cross-lingual representation alignment. Given its universality and simplicity, we anticipate that researchers could leverage the simplicity of CrossConST as a foundation to achieve new SOTA results in their own work. For future work, we will explore the effectiveness of CrossConST on more multilingual tasks, such as multilingual sentence embedding, multilingual word alignment, etc.

\section*{Limitations}

In this paper, we mainly focus on evaluating our approach on two English-centric corpora, IWSLT17 and PC32. Future research could consider more multilingual machine translation benchmarks with different number of languages and training samples and conduct experiments on more challenging training scenarios such as chain configurations where we have multiple bridge languages and different zero-shot distances.


\section*{Acknowledgements}

We would like to thank the anonymous reviewers for their insightful comments.

\bibliography{anthology,custom}
\bibliographystyle{acl_natbib}

\appendix

\section{Theoretical Discussion of CrossConST}\label{sec:crossconst_theory}

We first discuss how to derive the lower bound of $\mathcal{L}(\theta)$ as follows. Please note that we drop $\theta$ in the following proofs for the simplicity of the expression.

\begin{align*}
& \mathcal{L}(\theta) \\ 
& = \sum_{(\mathbf{x}, \mathbf{y}) \in \mathcal{S}} \log P(\mathbf{y} | \mathbf{x}) \\
& = \sum_{(\mathbf{x}, \mathbf{y}) \in \mathcal{S}} \log \sum_{\mathbf{z}} P(\mathbf{y} | \mathbf{x}, \mathbf{z}) P(\mathbf{z} | \mathbf{x}) \\
& \approx \sum_{(\mathbf{x}, \mathbf{y}) \in \mathcal{S}} \log \sum_{\mathbf{z}} P(\mathbf{y} | \mathbf{z}) P(\mathbf{z} | \mathbf{x}) \\
& = \sum_{(\mathbf{x}, \mathbf{y}) \in \mathcal{S}} \log \sum_{\mathbf{z}} P(\mathbf{z} | \mathbf{z}) \frac{P(\mathbf{y} | \mathbf{z}) P(\mathbf{z} | \mathbf{x})}{P(\mathbf{z} | \mathbf{z})} \\
& \ge \sum_{(\mathbf{x}, \mathbf{y}) \in \mathcal{S}} \sum_{\mathbf{z}} P(\mathbf{z} | \mathbf{z}) \log \frac{P(\mathbf{y} | \mathbf{z}) P(\mathbf{z} | \mathbf{x})}{P(\mathbf{z} | \mathbf{z})} \\
& = \sum_{(\mathbf{x}, \mathbf{y}) \in \mathcal{S}} \mathop{\mathbb{E}}_{\mathbf{z} \sim P(\mathbf{z} | \mathbf{z})} \log P(\mathbf{y} | \mathbf{z}) \\
& \ \ \ \ \ - \text{KL}(P(\mathbf{z} | \mathbf{z})\|P(\mathbf{z} | \mathbf{x})) \\
& := \bar{\mathcal{L}}(\theta)
\end{align*}

We then discuss how to derive the gap between $\mathcal{L}(\theta)$ and $\bar{\mathcal{L}}(\theta)$ as follows.

\begin{align*}
& \mathcal{L}(\theta) - \bar{\mathcal{L}}(\theta) \\
& = \sum_{(\mathbf{x}, \mathbf{y}) \in \mathcal{S}} \sum_{\mathbf{z}} P(\mathbf{z} | \mathbf{z}) \log \frac{P(\mathbf{z} | \mathbf{z})P(\mathbf{y} | \mathbf{x})}{P(\mathbf{y} | \mathbf{z}) P(\mathbf{z} | \mathbf{x})} \\
& = \sum_{(\mathbf{x}, \mathbf{y}) \in \mathcal{S}} \sum_{\mathbf{z}} P(\mathbf{z} | \mathbf{z}) \log \frac{P(\mathbf{z} | \mathbf{z}) P(\mathbf{y} | \mathbf{x}) P(\mathbf{z} | \mathbf{y})}{P(\mathbf{y} | \mathbf{z}) P(\mathbf{z} | \mathbf{x}) P(\mathbf{z} | \mathbf{y})} \\
& \approx \sum_{(\mathbf{x}, \mathbf{y}) \in \mathcal{S}} \sum_{\mathbf{z}} P(\mathbf{z} | \mathbf{z}) \log \frac{P(\mathbf{z} | \mathbf{z}) P(\mathbf{y} | \mathbf{x}) P(\mathbf{z} | \mathbf{y})}{P(\mathbf{y} | \mathbf{x}, \mathbf{z}) P(\mathbf{z} | \mathbf{x}) P(\mathbf{z} | \mathbf{y})} \\
& = \sum_{(\mathbf{x}, \mathbf{y}) \in \mathcal{S}} \sum_{\mathbf{z}} P(\mathbf{z} | \mathbf{z}) \log \frac{P(\mathbf{z} | \mathbf{z}) P(\mathbf{y} | \mathbf{x}) P(\mathbf{z} | \mathbf{y})}{P(\mathbf{y}, \mathbf{z} | \mathbf{x}) P(\mathbf{z} | \mathbf{y})} \\
& = \sum_{(\mathbf{x}, \mathbf{y}) \in \mathcal{S}} \sum_{\mathbf{z}} P(\mathbf{z} | \mathbf{z}) \log \frac{P(\mathbf{z} | \mathbf{z}) P(\mathbf{y} | \mathbf{x}) P(\mathbf{z} | \mathbf{y})}{P(\mathbf{z} | \mathbf{x}, \mathbf{y}) P(\mathbf{y} | \mathbf{x}) P(\mathbf{z} | \mathbf{y})} \\
& \approx \sum_{(\mathbf{x}, \mathbf{y}) \in \mathcal{S}} \sum_{\mathbf{z}} P(\mathbf{z} | \mathbf{z}) \log \frac{P(\mathbf{z} | \mathbf{z}) P(\mathbf{y} | \mathbf{x}) P(\mathbf{z} | \mathbf{y})}{P(\mathbf{z} | \mathbf{y}) P(\mathbf{y} | \mathbf{x}) P(\mathbf{z} | \mathbf{y})} \\
& = \sum_{(\mathbf{x}, \mathbf{y}) \in \mathcal{S}} \sum_{\mathbf{z}} P(\mathbf{z} | \mathbf{z}) \log \frac{P(\mathbf{z} | \mathbf{z})}{P(\mathbf{z} | \mathbf{y})} \\
& = \sum_{(\mathbf{x}, \mathbf{y}) \in \mathcal{S}} \text{KL}(P(\mathbf{z} | \mathbf{z}) \| P(\mathbf{z} | \mathbf{y})),
\end{align*}
where we utilize two approximations as follows:
\begin{equation}
P(\mathbf{y} | \mathbf{x}, \mathbf{z}) \approx P(\mathbf{y} | \mathbf{z})
\end{equation}
and
\begin{equation}
P(\mathbf{z} | \mathbf{x}, \mathbf{y}) \approx P(\mathbf{z} | \mathbf{y}).
\end{equation}

\section{Statistics of all training datasets}\label{sec:appendix_dataset}

\begin{table}[h]
\centering
\begin{tabular}{c | c | c | c} 
\texttt{en} $\leftrightarrow$ & \#sentences & \texttt{en} $\leftrightarrow$ & \#sentences \\
\hline
\hline
\texttt{de} & $446324$ & \texttt{nl} & $510580$ \\
\texttt{it} & $501278$ & \texttt{ro} & $477316$ \\
\hline
\end{tabular}
\caption{Statistics of IWSLT17 dataset. Each entry shows the total number of parallel sentence pairs for both directions. Note that \texttt{en} $\rightarrow$ and \texttt{en} $\leftarrow$ directions have the equal number of sentence pairs.}
\label{iwslt17_statics}
\end{table}

\begin{table}[h]
\centering
\begin{tabular}{c | c | c | c} 
\texttt{en} $\leftrightarrow$ & \#sentences & \texttt{en} $\leftrightarrow$ & \#sentences \\
\hline
\hline
\texttt{af} & $80616$ & \texttt{ja} & $4146998$ \\
\texttt{ar} & $2424336$ & \texttt{ka} & $400868$ \\
\texttt{be} & $51008$ & \texttt{kk} & $246622$ \\
\texttt{bg} & $6305372$ & \texttt{ko} & $2945682$ \\
\texttt{cs} & $1639292$ & \texttt{lt} & $4721996$ \\
\texttt{de} & $9420278$ & \texttt{lv} & $6261224$ \\
\texttt{el} & $2678292$ & \texttt{mn} & $61200$ \\
\texttt{eo} & $134972$ & \texttt{ms} & $3273034$ \\
\texttt{es} & $4228938$ & \texttt{mt} & $354488$ \\
\texttt{et} & $4579720$ & \texttt{my} & $57076$ \\
\texttt{fi} & $4113282$ & \texttt{ro} & $1550552$ \\
\texttt{fr} & $74445068$ & \texttt{ru} & $3686958$ \\
\texttt{gu} & $22792$ & \texttt{sr} & $269302$ \\
\texttt{he} & $664818$ & \texttt{tr} & $771426$ \\
\texttt{hi} & $2699732$ & \texttt{vi} & $6450690$ \\
\texttt{it} & $4144732$ & \texttt{zh} & $44771930$ \\
\hline
\end{tabular}
\caption{Statistics of PC32 dataset. Each entry shows the total number of parallel sentence pairs for both directions. Note that \texttt{en} $\rightarrow$ and \texttt{en} $\leftarrow$ directions have the equal number of sentence pairs.}
\label{pc32_statics}
\end{table}

\begin{table*}[h]
\centering
\begin{tabular}{c | c | c | c | c | c | c | c} 
\texttt{en} $\rightarrow$ & \#sentences & \texttt{en} $\leftarrow$ & \#sentences & \texttt{en} $\rightarrow$ & \#sentences & \texttt{en} $\leftarrow$ & \#sentences \\
\hline
\hline
\texttt{af} & $58723$ & \texttt{af} & $42429$ & \texttt{ja} & $2989787$ & \texttt{ja} & $2072284$ \\
\texttt{ar} & $1786139$ & \texttt{ar} & $1212160$ & \texttt{ka} & $281346$ & \texttt{ka} & $200434$ \\
\texttt{be} & $41052$ & \texttt{be} & $25504$ & \texttt{kk} & $132937$ & \texttt{kk} & $123309$ \\
\texttt{bg} & $5360004$ & \texttt{bg} & $3152631$ & \texttt{ko} & $2130540$ & \texttt{ko} & $1472841$ \\
\texttt{cs} & $1455275$ & \texttt{cs} & $819418$ & \texttt{lt} & $3545300$ & \texttt{lt} & $2359916$ \\
\texttt{de} & $8251292$ & \texttt{de} & $4707481$ & \texttt{lv} & $5179183$ & \texttt{lv} & $3130536$ \\
\texttt{el} & $2402732$ & \texttt{el} & $1333533$ & \texttt{mn} & $49882$ & \texttt{mn} & $30600$ \\
\texttt{eo} & $93519$ & \texttt{eo} & $67486$ & \texttt{ms} & $2268324$ & \texttt{ms} & $1636517$ \\
\texttt{es} & $3787101$ & \texttt{es} & $2111065$ & \texttt{mt} & $306122$ & \texttt{mt} & $177244$ \\
\texttt{et} & $3289592$ & \texttt{et} & $2289755$ & \texttt{my} & $48497$ & \texttt{my} & $28538$ \\
\texttt{fi} & $3571662$ & \texttt{fi} & $2054925$ & \texttt{ro} & $1359006$ & \texttt{ro} & $775197$ \\
\texttt{fr} & $63591612$ & \texttt{fr} & $37222318$ & \texttt{ru} & $2859034$ & \texttt{ru} & $1843417$ \\
\texttt{gu} & $11868$ & \texttt{gu} & $11395$ & \texttt{sr} & $229641$ & \texttt{sr} & $134651$ \\
\texttt{he} & $532895$ & \texttt{he} & $332357$ & \texttt{tr} & $660576$ & \texttt{tr} & $385713$ \\
\texttt{hi} & $1990436$ & \texttt{hi} & $1349767$ & \texttt{vi} & $4542508$ & \texttt{vi} & $3225345$ \\
\texttt{it} & $3733382$ & \texttt{it} & $2068077$ & \texttt{zh} & $37297105$ & \texttt{zh} & $22385733$ \\
\hline
\end{tabular}
\caption{Statistics of PC32 with RAS dataset. Each entry shows the total number of parallel sentence pairs for each direction.}
\label{pc32_ras_statics}
\end{table*}

\section{Details of Evaluation Results on IWSLT17}\label{sec:iwslt17_eval}

\begin{table*}[h]
\centering
\begin{tabular}{l | c c c c c c c c c c} 
\multicolumn{1}{c|}{Method} & \multicolumn{2}{c}{\texttt{de} - \texttt{it}} & \multicolumn{2}{c}{\texttt{de} - \texttt{nl}} & \multicolumn{2}{c}{\texttt{de} - \texttt{ro}} & \multicolumn{2}{c}{\texttt{it} - \texttt{ro}} & \multicolumn{2}{c}{\texttt{it} - \texttt{nl}} \\
& $\rightarrow$ & $\leftarrow$ & $\rightarrow$ & $\leftarrow$ & $\rightarrow$ & $\leftarrow$ & $\rightarrow$ & $\leftarrow$ & $\rightarrow$ & $\leftarrow$ \\
\hline
\hline
Pivot & 18.81 & 18.92 & 19.87 & 20.3 & 16.26 & 18.13 & 20.19 & 22.93 & 22.2 & 22.23 \\
\hdashline
OT \& AT & 18.17 & 18.18 & 20.17 & 20.27 & 16.12 & 17.52 & 20.14 & 23.77 & 21.07 & 21.22 \\
m-Transformer & 17.18 & 17.22 & 19.21 & 20.01 & 15.21 & 16.54 & 19.27 & 22.35 & 20.31 & 20.1 \\
\ \ + CrossConST & 18.6 & 18.79 & 20.41 & 20.22 & 15.9 & 18.06 & 21.02 & 23.31 & 21.88 & 21.77 \\
\hline
\\
\multicolumn{1}{c|}{Method} & \multicolumn{2}{c}{\texttt{nl} - \texttt{ro}} & \multicolumn{2}{c}{\texttt{en} - \texttt{de}} & \multicolumn{2}{c}{\texttt{en} - \texttt{it}} & \multicolumn{2}{c}{\texttt{en} - \texttt{nl}} & \multicolumn{2}{c}{\texttt{en} - \texttt{ro}} \\
& $\rightarrow$ & $\leftarrow$ & $\rightarrow$ & $\leftarrow$ & $\rightarrow$ & $\leftarrow$ & $\rightarrow$ & $\leftarrow$ & $\rightarrow$ & $\leftarrow$ \\
\hline
\hline
Pivot & 18.06 & 20.64 & - & - & - & - & - & - & - & - \\
\hdashline
OT \& AT & 17.81 & 19.51 & 24.87 & 28.67 & 35.29 & 37.61 & 31.04 & 33.03 & 26.17 & 32.45 \\
m-Transformer & 16.65 & 19.12 & 24.73 & 28.49 & 35.34 & 38.12 & 31.64 & 33.47 & 26.36 & 32.56 \\
\ \ + CrossConST & 18.21 & 20.38 & 24.7 & 28.87 & 35.02 & 38.18 & 31.75 & 33.16 & 26.65 & 32.66 \\
\end{tabular}
\caption{Performance on IWSLT17 supervised and zero-shot translation directions with the English-centric training dataset.}
\label{iwslt17-eval}
\end{table*}

\begin{table*}[h]
\centering
\begin{tabular}{l | c c c c c c c c c c} 
\multicolumn{1}{c|}{Method} & \multicolumn{2}{c}{\texttt{de} - \texttt{it}} & \multicolumn{2}{c}{\texttt{de} - \texttt{nl}} & \multicolumn{2}{c}{\texttt{de} - \texttt{ro}} & \multicolumn{2}{c}{\texttt{it} - \texttt{ro}} & \multicolumn{2}{c}{\texttt{it} - \texttt{nl}} \\
& $\rightarrow$ & $\leftarrow$ & $\rightarrow$ & $\leftarrow$ & $\rightarrow$ & $\leftarrow$ & $\rightarrow$ & $\leftarrow$ & $\rightarrow$ & $\leftarrow$ \\
\hline
\hline
m-Transformer & 18.55 & 18.88 & 20.5 & 20.56 & 16.06 & 17.93 & 20.47 & 23.42 & 22.06 & 21.66 \\
\ \ + CrossConST & 19.35 & 19.63 & 20.69 & 20.7 & 16.48 & 18.33 & 21.23 & 23.74 & 22.75 & 22.31 \\
\hline
\\
\multicolumn{1}{c|}{Method} & \multicolumn{2}{c}{\texttt{nl} - \texttt{ro}} & \multicolumn{2}{c}{\texttt{en} - \texttt{de}} & \multicolumn{2}{c}{\texttt{en} - \texttt{it}} & \multicolumn{2}{c}{\texttt{en} - \texttt{nl}} & \multicolumn{2}{c}{\texttt{en} - \texttt{ro}} \\
& $\rightarrow$ & $\leftarrow$ & $\rightarrow$ & $\leftarrow$ & $\rightarrow$ & $\leftarrow$ & $\rightarrow$ & $\leftarrow$ & $\rightarrow$ & $\leftarrow$ \\
\hline
\hline
m-Transformer & 17.79 & 19.28 & 25.24 & 29.1 & 35.42 & 38.32 & 31.09 & 33.32 & 26.95 & 33.3 \\
\ \ + CrossConST & 18.45 & 20.57 & 24.88 & 29.35 & 35.46 & 38.39 & 31.41 & 33.38 & 26.87 & 33.65 \\
\end{tabular}
\caption{Performance on IWSLT17 supervised and zero-shot translation directions with the English-centric and extra \texttt{de} $\leftrightarrow$ \texttt{it} training dataset.}
\label{iwslt17-extra-eval}
\end{table*}

\section{Details of Evaluation Results on OPUS-100}\label{sec:zeroshot}

\begin{table*}[h]
\centering
\begin{tabular}{c | c c c c c | c | c | c c c c c | c}
\hline
\multicolumn{7}{c|}{\textbf{m-Transformer}} & \multicolumn{7}{c}{\textbf{m-Transformer + CrossConST}} \\
\hline
 & \texttt{ar} & \texttt{zh} & \texttt{fr} & \texttt{de} & \texttt{ru} & Avg &  & \texttt{ar} & \texttt{zh} & \texttt{fr} & \texttt{de} & \texttt{ru} & Avg \\
\hline
\texttt{ar}$\rightarrow$ & - & 15.8 & 9.4 & 6.6 & 13.0 & 11.2 & \texttt{ar}$\rightarrow$ & - & 27.6 & 19.1 & 11.0 & 13.1 & 17.7 \\
\texttt{zh}$\rightarrow$ & 6.4 & - & 33.1 & 6.6 & 19.9 & 16.5 & \texttt{zh}$\rightarrow$ & 6.2 & - & 36.4 & 9.6 & 21.3 & 18.4 \\
\texttt{fr}$\rightarrow$ & 6.8 & 40.0 & - & 16.3 & 22.2 & 21.3 & \texttt{fr}$\rightarrow$ & 7.0 & 43.6 & - & 21.1 & 24.0 & 23.9 \\
\texttt{de}$\rightarrow$ & 4.2 & 16.5 & 18.6 & - & 13.2 & 13.1 & \texttt{de}$\rightarrow$ & 4.9 & 19.6 & 24.4 & - & 15.2 & 16.0 \\
\texttt{ru}$\rightarrow$ & 6.6 & 7.7 & 9.8 & 8.5 & - & 8.2 & \texttt{ru}$\rightarrow$ & 5.7 & 37.7 & 24.1 & 14.3 & - & 20.5 \\
\texttt{nl}$\rightarrow$ & 2.3 & 6.8 & 12.9 & 11.1 & 4.4 & 7.5 & \texttt{nl}$\rightarrow$ & 3.1 & 7.6 & 16.2 & 13.8 & 5.9 & 9.3 \\
\hline
Avg & 5.3 & 17.4 & 16.8 & 9.8 & 14.5 & \bf 12.75 & Avg & 5.4 & 27.2 & 24.0 & 14.0 & 15.9 & \bf 17.30 \\
\hline
\hline
\multicolumn{7}{c|}{\textbf{mRASP}} & \multicolumn{7}{c}{\textbf{mRASP + CrossConST}} \\
\hline
 & \texttt{ar} & \texttt{zh} & \texttt{fr} & \texttt{de} & \texttt{ru} & Avg &  & \texttt{ar} & \texttt{zh} & \texttt{fr} & \texttt{de} & \texttt{ru} & Avg \\
\hline
\texttt{ar}$\rightarrow$ & - & 22.6 & 10.6 & 7.7 & 13.7 & 13.7 & \texttt{ar}$\rightarrow$ & - & 26.2 & 16.0 & 11.0 & 13.4 & 16.7 \\
\texttt{zh}$\rightarrow$ & 7.1 & - & 35.2 & 9.4 & 21.6 & 18.3 & \texttt{zh}$\rightarrow$ & 6.7 & - & 37.3 & 11.6 & 22.9 & 19.6 \\
\texttt{fr}$\rightarrow$ & 7.4 & 41.9 & - & 18.7 & 24.1 & 23.0 & \texttt{fr}$\rightarrow$ & 7.8 & 43.9 & - & 21.6 & 24.9 & 24.6 \\
\texttt{de}$\rightarrow$ & 4.0 & 16.6 & 17.2 & - & 14.4 & 13.1 & \texttt{de}$\rightarrow$ & 4.9 & 19.6 & 24.3 & - & 15.3 & 16.0 \\
\texttt{ru}$\rightarrow$ & 7.2 & 33.6 & 11.8 & 9.3 & - & 15.5 & \texttt{ru}$\rightarrow$ & 6.9 & 38.8 & 24.0 & 13.9 & - & 20.9 \\
\texttt{nl}$\rightarrow$ & 2.4 & 5.9 & 13.6 & 10.4 & 3.7 & 7.2 & \texttt{nl}$\rightarrow$ & 3.3 & 7.5 & 15.9 & 13.6 & 5.7 & 9.2 \\
\hline
Avg & 5.6 & 24.1 & 17.7 & 11.1 & 15.5 & \bf 14.80 & Avg & 5.9 & 27.2 & 23.5 & 14.3 & 16.4 & \bf 17.48 \\
\hline
\hline
\multicolumn{7}{c|}{\textbf{Pivot (m-Transformer)}} & \multicolumn{7}{c}{\textbf{Pivot (mRASP)}} \\
\hline
 & \texttt{ar} & \texttt{zh} & \texttt{fr} & \texttt{de} & \texttt{ru} & Avg &  & \texttt{ar} & \texttt{zh} & \texttt{fr} & \texttt{de} & \texttt{ru} & Avg \\
\hline
\texttt{ar}$\rightarrow$ & - & 33.0 & 24.1 & 14.0 & 17.7 & 22.2 & \texttt{ar}$\rightarrow$ & - & 32.2 & 23.1 & 14.3 & 17.8 & 21.9 \\
\texttt{zh}$\rightarrow$ & 8.8 & - & 38.1 & 13.2 & 25.5 & 21.4 & \texttt{zh}$\rightarrow$ & 9.5 & - & 38.3 & 13.0 & 26.3 & 21.8 \\
\texttt{fr}$\rightarrow$ & 7.9 & 44.3 & - & 21.9 & 24.7 & 24.7 & \texttt{fr}$\rightarrow$ & 8.4 & 44.5 & - & 22.5 & 25.6 & 25.3 \\
\texttt{de}$\rightarrow$ & 5.2 & 21.4 & 25.7 & - & 16.2 & 17.1 & \texttt{de}$\rightarrow$ & 5.0 & 21.7 & 25.6  & - & 16.6 & 17.2 \\
\texttt{ru}$\rightarrow$ & 8.1 & 41.4 & 34.8 & 16.8 & - & 25.3 & \texttt{ru}$\rightarrow$ & 8.9 & 41.8 & 35.0 & 16.7 & - & 25.6 \\
\texttt{nl}$\rightarrow$ & 2.9 & 7.6 & 14.8 & 12.6 & 5.7 & 8.7 & \texttt{nl}$\rightarrow$ & 2.9 & 7.0 & 14.1 & 11.2 & 5.2 & 8.1 \\
\hline
Avg & 6.6 & 29.5 & 27.5 & 15.7 & 18.0 & \bf 19.46 & Avg & 6.9 & 29.4 & 27.2 & 15.5 & 18.3 & \bf 19.49 \\
\end{tabular}
\caption{Performance (de-tokenized BLEU using SacreBLEU) on OPUS100 zero-shot translation directions.}
\label{main_zeroshot_details}
\end{table*}

\section{Sentence Representation Visualization}\label{sec:visualization}

\begin{figure}[h]
\centering
\includegraphics[scale=0.45]{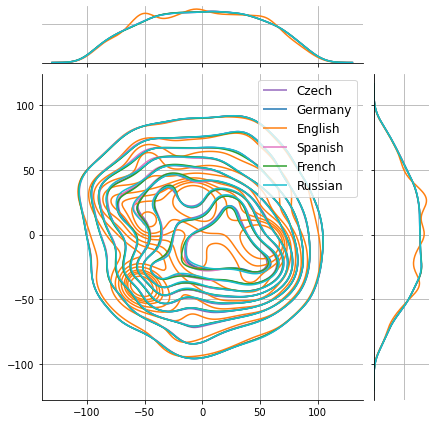}
\caption{Bivariate kernel density estimation plots of sentence representations based on m-Transformer without CrossConST.}
\label{fig:kde_baseline}
\end{figure}

\begin{figure}[h]
\centering
\includegraphics[scale=0.45]{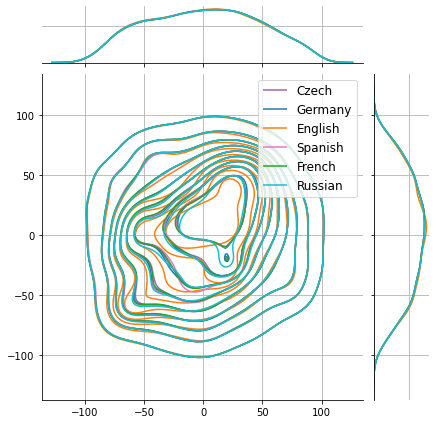}
\caption{Bivariate kernel density estimation plots of sentence representations based on m-Transformer with CrossConST.}
\label{fig:kde_crossconst}
\end{figure}

\end{document}